\documentclass[11pt]{article}

\usepackage[margin=1in]{geometry}
\usepackage{amsmath, amssymb, amsthm}
\usepackage{graphicx}
\usepackage{booktabs}
\usepackage{hyperref}
\usepackage{natbib}
\usepackage{microtype}
\usepackage{enumitem}
\usepackage{multirow}
\usepackage{hyperref}

\title{Directional Optimization Asymmetry in Transformers:\\
A Synthetic Stress Test}
\date{}
\author{
  Mihir Sahasrabudhe\thanks{Email: \texttt{[mihirss2@illinois.edu]}.}\\
}

\begin{document}

\maketitle

\begin{abstract}

Transformers are theoretically reversal-invariant: their function class does not prefer left-to-right over right-to-left mappings. Yet empirical studies on natural language repeatedly report a “reversal curse,” and recent work on temporal asymmetry in LLMs suggests that real-world corpora carry their own arrow of time. This leaves an unresolved question: do directional failures stem from linguistic statistics, or from the architecture itself?
We cut through this ambiguity with a fully synthetic, entropy-controlled benchmark designed as a clean-room stress test for directional learning. Using random string mappings with tunable branching factor K, we construct forward tasks with zero conditional entropy and inverse tasks with analytically determined entropy floors. Excess loss above these floors reveals that even scratch-trained GPT-2 models exhibit a strong, reproducible directional optimization gap (e.g., 1.16 nats at K=5), far larger than that of an MLP trained on the same data. Pre-trained initializations shift optimization behavior but do not eliminate this gap, while LoRA encounters a sharp capacity wall on high-entropy inverse mappings.
Together, these results isolate a minimal, semantics-free signature of directional friction intrinsic to causal Transformer training—one that persists even when linguistic priors, token frequencies, and corpus-level temporal asymmetries are removed. Our benchmark provides a controlled instrument for dissecting directional biases in modern sequence models and motivates deeper mechanistic study of why inversion remains fundamentally harder for Transformers.

\end{abstract}

\section{Introduction}

While Transformer-based Large Language Models (LLMs) excel at sequence modeling, they exhibit a fundamental directional asymmetry. Recent studies report that models trained on causal statements ($A \to B$) often fail to infer the diagnostic inverse ($B \to A$), a phenomenon termed the ``reversal curse'' \citep{berglund2024reversalcursellmstrained}. This raises a critical question: is this asymmetry an artifact of the training data, or an intrinsic inductive bias of the architecture itself?

Current analyses relying on natural language (e.g., ``$A$ is the parent of $B$'') are inherently confounded by three factors:
\begin{enumerate}[label=\arabic*., nosep]
    \item \textbf{Semantic Priors:} Causal relationships often appear more frequently than diagnostic ones in training corpora.
    \item \textbf{Linguistic Structure:} Syntax and grammar impose directional dependencies (e.g., Subject-Verb-Object) that favor forward prediction.
    \item \textbf{Token Statistics:} Entity frequencies and co-occurrence statistics are rarely symmetric.
\end{enumerate}
Consequently, it is difficult to disentangle whether the Reversal Curse arises from data distribution or from the autoregressive factorization mechanism.

\paragraph{The Arrow of Complexity.}
This directional puzzle is further complicated by inherent differences in computational complexity. As noted in recent work on the ``Arrows of Time'' in language models \citep{papadopoulos2024arrowstimelargelanguage}, forward generation often resembles a deterministic collapse of state space (analogous to multiplication), whereas inverse inference requires expanding state space to recover multiple potential inputs (analogous to factorization). In natural language, these entropic differences are inextricably linked to semantics. To isolate the architectural contribution to the Reversal Curse, we require a setting where this forward--inverse complexity asymmetry is explicitly tunable.

\paragraph{Goal: A Synthetic ``Clean Room''.}
We introduce a controlled benchmark to measure \emph{directional training efficiency} in the absence of linguistic or statistical confounds. We construct a dataset of random string mappings where the topology is strictly controlled by a branching factor $K$.
\begin{itemize}[nosep]
    \item \textbf{Forward ($A \to B$):} Deterministic mapping ($H=0$), mimicking low-entropy causal processes.
    \item \textbf{Backward ($B \to A$):} Probabilistic one-to-many mapping ($H=\ln K$), mimicking high-entropy inverse problems.
\end{itemize}
This design creates a mathematically precise analogue of the complexity asymmetry described in prior work, stripped of all semantic priors.

\paragraph{Approach.}
Within this framework, we benchmark the optimization dynamics of Causal Transformers (trained from scratch and pre-trained) against non-causal Multilayer Perceptrons (MLPs) and Low-Rank Adaptation (LoRA) methods \citep{hu2021loralowrankadaptationlarge}. Crucially, because the information-theoretic floor of each task is known exactly, we report \emph{Excess Loss}---the divergence of the model from the theoretical minimum. This metric allows us to rigorously decouple the inherent thermodynamic difficulty of the inverse task from the structural inefficiencies of the architecture. \\
Code can be found at: \url{https://github.com/mihirs-0/synass}

\section{A Synthetic ``Clean Room'' for Directional Analysis}

To measure directional training behavior in the absence of linguistic or corpus-level confounds, we construct a minimal, fully synthetic environment. In this setting, both the data topology and the information-theoretic difficulty of the forward and inverse tasks are exactly known. This ``Clean Room'' isolates architectural and optimization effects from the semantic and statistical artifacts inherent to natural language.

\subsection{Data Construction}

We define a uniform alphabet $\Sigma = \{\texttt{a--z, 0--9}\}$ of size $|\Sigma|=36$. We sample fixed-length strings from $\Sigma^L$ (with $L=8$), drawn i.i.d. from a uniform distribution. By design, no token, substring, or structural pattern appears with higher-than-random frequency.

We construct mappings using a controlled branching factor $K$:
\begin{itemize}[nosep]
    \item For each unique target $B_i$, we sample $K$ distinct inputs $\{A_{i,1}, \dots, A_{i,K}\}$.
    \item Uniqueness constraint: Each $A_{i,j}$ appears exactly once globally.
    \item Support constraint: Each $B_i$ appears exactly $K$ times.
\end{itemize}

This construction yields two distinct tasks:

\paragraph{Forward ($A \to B$): Deterministic.}
The mapping $A_{i,j} \mapsto B_i$ is a function. The conditional entropy is $H(B \mid A) = 0$.

\paragraph{Inverse ($B \to A$): Probabilistic.}
The mapping $B_i \mapsto \{A_{i,j}\}$ is one-to-many. The optimal policy is a uniform distribution over the $K$ pre-images, yielding an entropy floor $H(A \mid B) = \log K$.

\subsection{Directional Neutrality}

A common concern with many-to-one synthetic datasets is that target duplication ($B_i$) might induce a hidden directional bias. Our construction avoids this for three reasons:

\begin{enumerate}[nosep]
\item \textbf{Unique Forward Inputs.} Every $A_{i,j}$ is sampled uniquely. There are no statistical shortcuts or frequency artifacts that privilege the forward direction.
\item \textbf{Necessary Inverse Support.} The repetition of $B_i$ is mathematically required to define the inverse conditional distribution. Without observing all pairs $(B_i, A_{i,j})$, the model cannot learn the full support of the inverse mapping.
\item \textbf{Empirical Symmetry at $K{=}1$.} In our scratch baseline (see Section~\ref{sec:results}), Transformers trained on the bijective case $K{=}1$ exhibit matched forward and reverse convergence after accounting for the entropy floor ($H(B\mid A)=H(A\mid B)=0$). This symmetry at $K{=}1$ confirms that the dataset itself does not impose a directional preference; the asymmetries observed for $K{>}1$ emerge only once the inverse task becomes many-to-one.
\end{enumerate}

Thus, the dataset is \emph{direction-neutral}: any efficiency gap during training must originate from the model architecture or optimization dynamics, not from artifacts of the synthetic mapping.

\subsection{Symmetric Prompting and Loss}

To ensure architectural symmetry, we use identical prompting structures and loss masking for both directions:
\begin{align*}
    \text{Forward:} & \quad \texttt{x: } A \texttt{ y: } B \\
    \text{Inverse:} & \quad \texttt{x: } B \texttt{ y: } A
\end{align*}
All prompt tokens (including delimiters) are masked ($\text{label}=-100$). The loss is computed exclusively over the target span. This ensures that the objective function is strictly:
\begin{equation}
    \mathcal{L} = -\log p_\theta(\text{target} \mid \text{prompt}),
\end{equation}
maintaining an identical conditioning structure for both tasks.

\subsection{Excess Loss as a Metric}

Because the data topology is fully controlled, the theoretical minimum achievable loss ($\mathcal{L}_{\min}$) is known analytically:
\begin{equation}
    \mathcal{L}_{\min} =
    \begin{cases}
        0 & \text{for } A \to B \\
        \log K & \text{for } B \to A
    \end{cases}
\end{equation}
We therefore report \textbf{Excess Loss}:
\begin{equation}
    \mathcal{L}_{\text{excess}} = \mathcal{L}_{\text{observed}} - \mathcal{L}_{\text{min}}.
\end{equation}
This metric isolates pure optimization inefficiency, decoupling the model's performance from the inherent thermodynamic difficulty of the inverse task.

\subsection{The ``Wind Tunnel'' Justification}

This setup functions as a computational wind tunnel. By ensuring that:
\begin{itemize}[nosep]
    \item All surface statistics are uniform,
    \item No semantic priors exist, and
    \item Supervision is structurally symmetric,
\end{itemize}
we guarantee that the sole asymmetry between $A \to B$ and $B \to A$ is the entropy difference induced by the mapping topology. Consequently, any residual directional gap in $\mathcal{L}_{\text{excess}}$ is strictly attributable to architectural inductive biases, optimization pathologies, or capacity limitations---not to the data.

\subsection{Contributions}

This work makes three practical contributions:
\begin{enumerate}[nosep]
    \item We introduce a synthetic benchmark for measuring directional training asymmetry in sequence models, based on random string mappings with tunable conditional entropy and analytically computable loss floors.
    \item We define and apply a normalized metric, \emph{Excess Loss}, to compare deterministic and probabilistic tasks across architectures, including Transformers trained from scratch, pre-trained Transformers, low-rank adaptation (LoRA), and non-causal MLPs.
    \item We provide empirical baselines that illustrate: (a) Transformers trained from scratch exhibit a larger directional gap than non-causal MLPs on the same task; (b) pre-trained weights are less efficient than random initialization on this synthetic task in the forward direction; and (c) LoRA struggles to match full fine-tuning on high-entropy inverse mappings at the examined scale.
\end{enumerate}

We intentionally frame these results conservatively: the benchmark is meant as an instrument for measuring directional behavior, not as a final explanation of natural-language reversal phenomena.

\section{Benchmark Design}

\subsection{Random String Mappings with Controlled Branching Factor}

Let $\Sigma$ denote an alphabet of size $|\Sigma| = 36$ (lowercase letters and digits). We consider strings of length $L = 8$:
\begin{equation}
    \mathcal{S} = \Sigma^L.
\end{equation}
We construct a dataset $\mathcal{D}$ of pairs $(A, B)$, where $A, B \in \mathcal{S}$, subject to a constraint on the mapping topology controlled by a branching factor $K$.

\paragraph{Case $K = 1$ (bijective mapping).}
We sample $n_{\text{pairs}}$ distinct strings for the $A$ side and $n_{\text{pairs}}$ distinct strings for the $B$ side, then shuffle $B$ and pair them. Each $A$ maps to a unique $B$ and vice versa:
\begin{equation}
    f : A \mapsto B, \quad f^{-1} : B \mapsto A,
\end{equation}
with no collisions.

\paragraph{Case $K > 1$ (many-to-one mapping).}
We instead sample $n_{\text{targets}} = n_{\text{pairs}} / K$ distinct target strings for $B$. For each $B$, we sample $K$ distinct $A$ strings and create pairs $(A, B)$. Each $B$ thus has $K$ pre-images, while each $A$ has a unique $B$.

In all cases, strings are drawn uniformly at random from $\mathcal{S}$, and the mapping structure is independent of any human notion of meaning.

\subsection{Forward and Inverse Tasks}

Given a dataset $\mathcal{D}$, we define two tasks:
\begin{description}[nosep]
    \item[Forward ($A \to B$):] Given $A$, predict $B$. This mapping is deterministic by construction, with one correct $B$ for each $A$.
    \item[Backward ($B \to A$):] Given $B$, predict $A$. For $K > 1$, there are $K$ valid $A$ values associated with each $B$, distributed uniformly by design.
\end{description}

We treat each string prediction task as sequence modeling at the character or token level, depending on the architecture.

\subsection{Information-Theoretic Floors}

Let the training objective be the standard token-level cross-entropy loss over a dataset of input-output pairs $(x, y)$:
\begin{equation}
    \mathcal{L}(\theta) = - \mathbb{E}_{(x, y) \sim \mathcal{D}} \left[ \sum_{t} \log p_{\theta}(y_t \mid y_{<t}, x) \right].
\end{equation}

In our setup, the conditional distribution $p^*(Y \mid X)$ implied by the data is simple:
\begin{itemize}[nosep]
    \item For the forward task, $A \to B$, the mapping is deterministic. An ideal model reproducing the mapping exactly would have conditional entropy
    \begin{equation}
        H(B \mid A) = 0,
    \end{equation}
    yielding an ideal loss floor $\mathcal{L}_{\min}^{A \to B} = 0$ (up to tokenization details).
    \item For the backward task, $B \to A$, when each $B$ has $K$ associated $A$ values, an optimal model that matches the uniform conditional distribution over these pre-images satisfies
    \begin{equation}
        H(A \mid B) = \log K,
    \end{equation}
    yielding an ideal loss floor $\mathcal{L}_{\min}^{B \to A} = \log K$ in nats, again abstracting away tokenization issues.
\end{itemize}

These quantities provide a reference for how difficult each task is \emph{in principle}, independent of the modeling architecture.

\subsection{Excess Loss}

To compare models and directions on equal terms, we define the \emph{Excess Loss} as the gap between the observed loss and the corresponding information-theoretic floor:
\begin{equation}
    \mathcal{L}_{\text{excess}} = \mathcal{L}_{\text{obs}} - \mathcal{L}_{\min}.
\end{equation}

For $A \to B$, we have $\mathcal{L}_{\min}^{A \to B} = 0$, so $\mathcal{L}_{\text{excess}}^{A \to B}$ coincides with the observed loss. For $B \to A$ with branching factor $K$, we use $\mathcal{L}_{\min}^{B \to A} = \log K$.

We define the \emph{directional gap} for a given architecture and configuration as
\begin{equation}
    \Delta_{\text{dir}} = \mathcal{L}_{\text{excess}}^{B \to A} - \mathcal{L}_{\text{excess}}^{A \to B}.
\end{equation}
A positive $\Delta_{\text{dir}}$ indicates higher inefficiency in the inverse direction.

\section{Architectures and Training Setup}

We report results for three Transformer regimes and one non-causal baseline.

\subsection{Transformer: GPT-2 Small}

For the sequence model, we use the GPT-2 Small architecture as implemented in the Hugging Face Transformers library. In all experiments, the maximum sequence length is set large enough to cover a simple prompt plus the target string (e.g., a short prefix such as ``x: A y:'' followed by $B$), and we apply standard causal masking.

We consider:
\begin{itemize}[nosep]
    \item \textbf{Scratch:} Random initialization from the GPT-2 Small configuration, with no pretraining.
    \item \textbf{Fine-tuning:} Initialization from pre-trained GPT-2 Small weights, then training on the synthetic mapping tasks.
    \item \textbf{Fine-tuning with regularization:} As above, but with increased dropout and weight decay; we treat this as a robustness variant rather than a separate contribution.
    \item \textbf{LoRA:} Low-Rank Adaptation applied to the pre-trained GPT-2 Small model, with various ranks $r$.
\end{itemize}

Unless otherwise noted, we train for $20$ epochs with batch size $64$ and learning rate $10^{-4}$ using AdamW with standard settings and linear warmup. We use $n_{\text{pairs}} = 40{,}000$ for each value of $K \in \{1, 5, 8\}$, and strings of length $L = 8$.

\subsection{Non-Causal Baseline: MLP}

To obtain a non-sequential baseline, we construct a simple character-level MLP:

\begin{itemize}[nosep]
    \item \textbf{Input:} Integer-encoded characters of the source string (length $L=8$).
    \item \textbf{Embedding:} Each character index is mapped to a $d_{\text{emb}} = 64$-dimensional embedding. These are concatenated into a single flat vector of dimension $L \cdot d_{\text{emb}}$.
    \item \textbf{Hidden Layers:} The flat vector is passed through a stack of $n_{\text{layers}} = 4$ fully connected blocks. Each block consists of a linear layer with hidden dimension $d_{\text{hidden}} = 512$, a ReLU activation, and Layer Normalization.
    \item \textbf{Output:} A final linear layer projects the hidden state to $L \cdot |\Sigma|$ logits, which are reshaped into $[L, |\Sigma|]$ to compute the token-level cross-entropy loss.
\end{itemize}

We train the MLP for $50$ epochs with a batch size of $256$ using the AdamW optimizer. We use a learning rate of $10^{-3}$ and set weight decay to $0$.

\section{Results}
\label{sec:results}

We report results in terms of \emph{excess loss} 
$\mathcal{L}_{\text{excess}} = \mathcal{L}_{\text{observed}} - \mathcal{L}_{\text{min}}$,
where $\mathcal{L}_{\text{min}} = 0$ for $A \to B$ and 
$\mathcal{L}_{\text{min}} = \log K$ for $B \to A$.
All experiments use $40{,}000$ pairs and sequence length $L=8$.

\subsection{Scratch Transformers vs. MLPs}

Table~\ref{tab:scratch_vs_mlp} compares GPT-2 (scratch) against the MLP baseline
for $K \in \{1,5,8\}$.
For $K=1$, the conditional entropy is zero in both directions and both models
exhibit nearly identical excess loss in $A \to B$ and $B \to A$.
As $K$ increases, the inverse task becomes higher-entropy, and both models incur
higher excess loss in the $B \to A$ direction.

However, the \emph{directional}
gap is substantially larger for the Transformer.
For example, at $K=5$, the scratch GPT-2 exhibits a gap of
$\approx 1.16$ nats between forward and inverse excess loss, whereas the MLP
shows a gap of only $\approx 0.22$ nats on the same data.
This pattern is consistent at $K=8$, where the Transformer gap remains
$\approx 0.90$ nats while the MLP gap is $\approx 0.11$ nats.

\begin{table}[t]
    \centering
    \renewcommand{\arraystretch}{1.2} % Adds vertical breathing room
    \setlength{\tabcolsep}{5pt}       % Adjusts column spacing to fit width
    
    \caption{\textbf{The Causal Tax.} Excess Loss (in nats) for the Scratch Transformer vs.\ the non-causal MLP baseline. While both models incur a small entropic penalty on the inverse task (due to the one-to-many mapping), the Transformer exhibits a massive architectural gap ($\Delta$) that is absent in the MLP.}
    \label{tab:scratch_vs_mlp}
    
    \begin{tabular}{cccccccc}
        \toprule
        & \multicolumn{3}{c}{\textbf{Transformer (Scratch)}} & & \multicolumn{3}{c}{\textbf{MLP (Baseline)}} \\
        \cmidrule(lr){2-4} \cmidrule(lr){6-8}
        $K$ & Forward & Inverse & \textbf{Gap ($\Delta$)} & & Forward & Inverse & Gap ($\Delta$) \\
        \midrule
        1 & 3.60 & 3.60 & 0.00 & & 0.48 & 0.50 & 0.01 \\
        \addlinespace[0.3em]
        5 & 0.91 & 2.07 & \textbf{1.16} & & 0.46 & 0.69 & 0.22 \\
        \addlinespace[0.3em]
        8 & 0.67 & 1.57 & \textbf{0.90} & & 0.42 & 0.53 & 0.11 \\
        \bottomrule
    \end{tabular}
\end{table}
These results indicate that some directional inefficiency is inherent to the
higher-entropy inverse task, but that the Transformer architecture amplifies
this effect relative to a non-causal MLP trained on the same mappings.

\subsection{Effect of Pre-training}

Table~\ref{tab:pretrain} summarizes excess loss for GPT-2 initialized from
pre-trained weights, with and without additional regularization.
For $K \in \{5,8\}$, the scratch model achieves lower forward excess loss
than the pre-trained variants (e.g., $0.91$ vs.\ $1.41$ at $K=5$),
suggesting that pre-trained weights are less plastic on this synthetic task
than random initialization.

The directional gap in excess loss for the pre-trained models is modest:
for $K=5$, the difference between $A \to B$ and $B \to A$ excess is
on the order of $0.06$--$0.09$ nats, and at $K=8$ the inverse task
sometimes attains slightly \emph{lower} excess loss than the forward task.
Within this benchmark, pre-training therefore appears to add a general
optimization penalty on arbitrary mappings, but does not introduce a strong
additional directional asymmetry beyond that seen from scratch.

\begin{table}[t]
    \centering
    \caption{Excess loss for GPT-2 pre-trained models on World~A.
    ``FT'' denotes full fine-tuning; ``FT-Reg'' adds higher weight decay and dropout.}
    \label{tab:pretrain}
    \begin{tabular}{lccccc}
        \toprule
        Regime & $K$ & Direction & Floor & Excess & Total Loss \\
        \midrule
        FT       & 1 & $A \to B$ & 0.00 & 3.03 & 3.03 \\
                 & 1 & $B \to A$ & 0.00 & 3.04 & 3.04 \\
        \midrule
        FT       & 5 & $A \to B$ & 0.00   & 1.41 & 1.41 \\
                 & 5 & $B \to A$ & 1.61   & 1.47 & 3.08 \\
        FT-Reg   & 5 & $A \to B$ & 0.00   & 1.37 & 1.37 \\
                 & 5 & $B \to A$ & 1.61   & 1.40 & 3.01 \\
        \midrule
        FT       & 8 & $A \to B$ & 0.00   & 1.17 & 1.17 \\
                 & 8 & $B \to A$ & 2.08   & 0.99 & 3.07 \\
        FT-Reg   & 8 & $A \to B$ & 0.00   & 1.14 & 1.14 \\
                 & 8 & $B \to A$ & 2.08   & 0.91 & 2.99 \\
        \bottomrule
    \end{tabular}
\end{table}

\subsection{LoRA Capacity under High-Entropy Mappings}

Table~\ref{tab:lora_collapse} reports excess loss for LoRA with ranks
$r \in \{8, 64, 256\}$.
Across all $K>1$, LoRA exhibits large excess loss in both directions and
does not approach the performance of either scratch or fully fine-tuned models.
For example, at $K=5$ the best LoRA setting ($r=256$) attains an inverse
excess loss of $3.15$ nats, compared to $2.07$ (scratch) and $1.47$ (full FT).

Directional gaps for LoRA are relatively small compared to the overall
excess loss, and the dominant effect in this benchmark is a capacity
limitation: low-rank adaptation fails to efficiently memorize these
high-entropy mappings at the given scale.

\begin{table}[t]
    \centering
    % This creates more vertical space between all rows
    \renewcommand{\arraystretch}{1.2} 
    
    \caption{\textbf{The LoRA Capacity Wall.} Excess Loss (in nats) for Low-Rank Adaptation on the synthetic benchmark. While LoRA achieves moderate success on the deterministic forward task ($A \to B$), it suffers from high excess loss on the high-entropy inverse task ($B \to A$), even at rank $r=256$.
    \emph{Note:} Theoretical floors are $0.00$ for Forward and $\ln K$ for Inverse.}
    \label{tab:lora_collapse}
    
    \begin{tabular}{ccccccc}
    \addlinespace
        \toprule
        & & \multicolumn{2}{c}{\textbf{Forward} ($A \to B$)} & & \multicolumn{2}{c}{\textbf{Inverse} ($B \to A$)} \\
        \cmidrule(lr){3-4} \cmidrule(lr){6-7}
        $K$ & Rank ($r$) & Excess & Loss & & \textbf{Excess} & Loss \\
        \midrule
        \multirow{3}{*}{5}
          & 8   & 4.97 & 4.97 & & 3.45 & 5.06 \\
          & 64  & 2.03 & 2.03 & & 3.24 & 4.85 \\
          & 256 & \textbf{1.82} & 1.82 & & \textbf{3.15} & 4.75 \\
        
        % This adds the specific gap you wanted
        \addlinespace[0.5em] 
        \midrule
        \addlinespace[0.5em]
        
        \multirow{3}{*}{8}
          & 8   & 4.85 & 4.85 & & 2.98 & 5.06 \\
          & 64  & 1.66 & 1.66 & & 2.77 & 4.85 \\
          & 256 & \textbf{1.60} & 1.60 & & \textbf{2.67} & 4.75 \\
        \bottomrule
    \end{tabular}
\end{table}

\begin{figure}[t]
    \centering
    \includegraphics[width=1.07\linewidth]{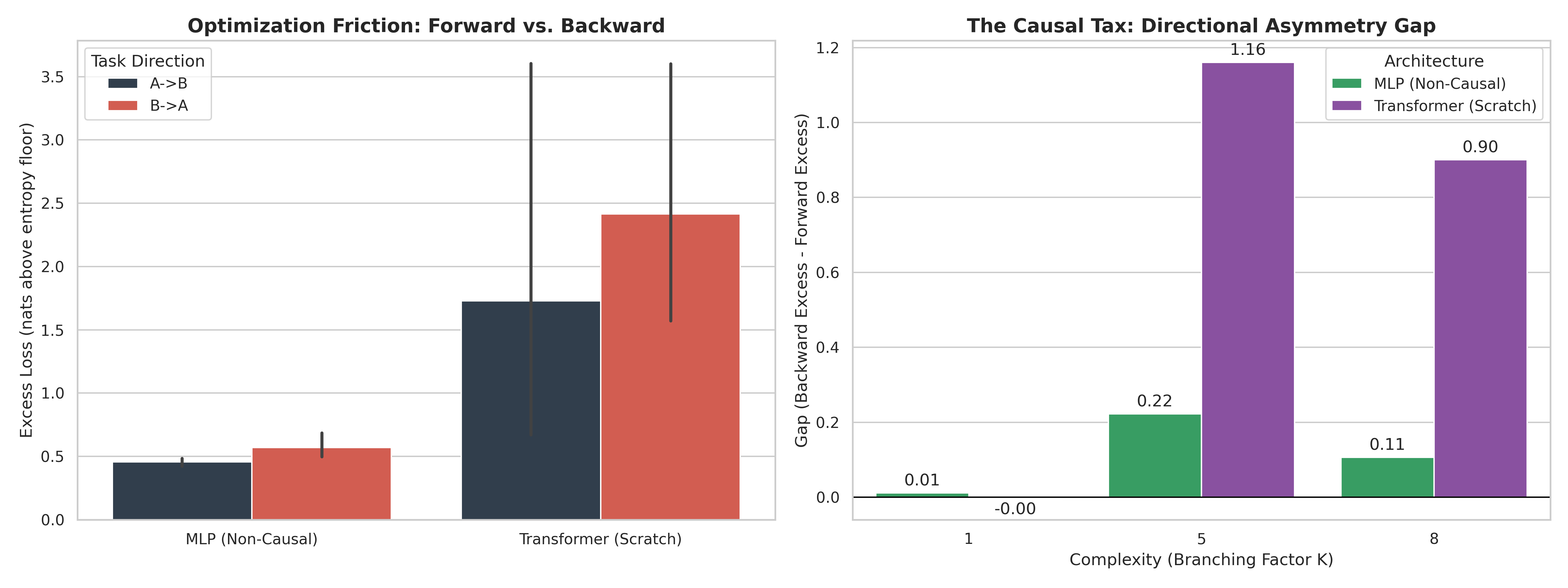}
    \caption{
        \textbf{Directional Optimization Friction.} 
        Excess loss (train loss minus entropy floor) for forward ($A\!\to\!B$) and inverse ($B\!\to\!A$) tasks across architectures and branching factors.
        Left: MLP vs.\ GPT-2 (causal LM, scratch initialization) showing higher excess loss for the inverse direction.
        Right: Directional asymmetry gap (inverse minus forward) as a function of branching factor $K$. 
        GPT-2 exhibits a substantially larger directional gap than the MLP on the same synthetic data.
    }
    \label{fig:directional_friction}
\end{figure}

\subsection{Summary}

Across all conditions, the synthetic benchmark reveals:

\begin{itemize}[nosep]
    \item A consistent directional gap for scratch-trained Transformers that grows with $K$ and exceeds the gap observed in MLPs on the same data.
    \item A general optimization penalty for pre-trained models on this synthetic task, with only mild additional directional effects.
    \item A pronounced capacity limitation for low-rank adaptation (LoRA), which fails to match full fine-tuning on high-entropy inverse mappings.
\end{itemize}

\subsection{Interpretation of Results}

Across all model classes and branching factors, the synthetic benchmark reveals three stable patterns.

\paragraph{(1) Scratch-trained Transformers exhibit a robust directional gap.}

For both $K=5$ and $K=8$, the scratch GPT-2 model consistently incurs substantially higher excess loss in the inverse ($B \to A$) direction than in the forward ($A \to B$) direction. The gap remains large (1.16 nats at $K=5$, 0.90 nats at $K=8$) and does not disappear with increased data.

Because the dataset is direction-neutral by construction, this gap reflects
optimization behavior induced by the causal Transformer architecture, not data artifacts.

We refrain from attributing causality beyond this: the metric indicates reduced efficiency in learning high-entropy inverse mappings relative to their deterministic counterparts.

\paragraph{(2) Pre-training alters efficiency but does not introduce strong additional asymmetry.}

Pre-trained GPT-2:
	•	Raises excess loss on the forward task (e.g., $1.41$ vs.\ $0.91$ at $K=5$).
	•	Lowers excess loss on the inverse task (e.g., $1.47$ vs.\ $2.07$ at $K=5$).
	•	Produces only modest directional differences (often $< 0.1$ nats).

Thus, pre-training mainly introduces a global optimization shift—sometimes helpful, sometimes harmful—rather than amplifying the directional gap. We interpret this conservatively as interference between pre-trained structure and arbitrary synthetic mappings, not as evidence of any directional bias induced by pre-training itself.

\paragraph{(3) LoRA is strongly capacity-limited on this workload.}

Across all ranks:
	•	Excess losses remain substantially higher than full fine-tuning.
	•	Directional gaps are small compared to the overall error.
	•	The dominant effect is the absolute inability to fit the mapping, rather than direction-specific behavior.

In this benchmark, LoRA behaves primarily as a low-capacity memorization mechanism rather than as an efficient optimizer under high-entropy inverse tasks. We do not generalize this beyond the present setup; we simply treat it as an empirical boundary condition for low-rank adapters on this synthetic load.

\paragraph{(4) MLPs show reduced directional sensitivity.}

The MLP is able to learn both directions with significantly lower excess loss and far smaller directional gaps (e.g., 0.22 nats at $K=5$).
This supports the interpretation that much of the observed Transformer gap is specific to causal, left-to-right factorization rather than the inherent difficulty of the inverse task itself.

\paragraph{Overall.}

The differences between models are stable and reproducible:
	•	Transformer (scratch): large directional gap
	•	Transformer (pre-trained): modest directionality; increased forward excess
	•	LoRA: dominated by capacity limits
	•	MLP: small directional gap

The benchmark therefore provides a simple way to quantify how various architectures respond to controlled increases in inverse-mapping entropy, without committing to any specific mechanistic explanation.

\section{Discussion}

\subsection{What the Benchmark Does and Does Not Claim}

The benchmark introduced here is deliberately narrow. Its purpose is not to generalize about natural language or to diagnose failure modes of real-world LLM behavior. Instead, it serves as a controlled instrument for studying directional learning under fully specified and entropy-matched conditions. To avoid misinterpretation, we highlight explicit boundaries:

\begin{itemize}[nosep]
\item \emph{Not a linguistic claim:} The benchmark does not attempt to explain why reversal failures occur in natural language tasks, where semantics, morphology, and token frequency play a dominant role.
\item \emph{Not a representational claim:} We do not argue that Transformers are incapable of learning inverse mappings. All models, including causal Transformers, can represent one-to-many conditional distributions in principle.
\item \emph{Not a universality claim:} The results here do not extend to arbitrary data modalities, large-scale training, or tasks with structure substantially richer than uniform random strings.
\end{itemize}

What the benchmark \emph{does} offer is a precise way to isolate the contribution of architecture and optimization dynamics when forward and inverse tasks differ only in their conditional entropy. The data distribution is symmetric; the mapping support is fully uniform; the token frequencies are flat. The only systematic asymmetry lies in $H(B \mid A) = 0$ versus $H(A \mid B) = \log K$, which is fully known and analytically controlled.

Within this framework, we observe three reproducible patterns:

\begin{enumerate}[nosep]
\item \textbf{Directional inefficiency in scratch Transformers.}
Scratch-trained GPT-2 consistently exhibits a larger excess loss on the inverse task than on the forward task, even though the data distribution itself does not prefer either direction. This suggests that the training dynamics of causal Transformers impose additional friction on high-entropy inverse mappings beyond what is required by the information-theoretic floor.
\item \textbf{Substantially smaller directional gap in MLPs.}  
When trained on the same data, MLPs show only a small directional excess loss gap. Since the MLP does not rely on autoregressive factorization or causal masking, this contrast indicates that the observed Transformer asymmetry is not inherent to the inverse task itself but is mediated by architectural or optimization choices.

\item \textbf{Interactions with pre-training and low-rank updates.}  
Pre-trained models and LoRA-adapted models display behavior that differs from both scratch Transformers and MLPs. Pre-training appears to shift the optimization landscape in nontrivial ways, while LoRA encounters hard capacity limitations under these synthetic, high-entropy mappings. These effects illustrate how initialization and parameterization interact with directional learning—not in natural language but in this controlled synthetic regime.
\end{enumerate}

\subsection{Implications and Interpretive Boundaries}

The controlled environment enables a type of attribution that is difficult to obtain from natural language experiments. Since the data support is symmetric and entropy is explicitly set, any residual directional gap must arise from:

\begin{itemize}[nosep]
\item the causal architecture,
\item the autoregressive training objective,
\item optimization dynamics under this parameterization,
\item initialization-induced inductive biases (in the case of pre-training),
\item or capacity constraints (in the case of LoRA).
\end{itemize}

The benchmark thus functions as a kind of ``wind tunnel’’ for assessing how architectures behave when learning forward versus inverse mappings without the confounds of linguistic structure. It does not diagnose the ultimate cause of asymmetry; it only demonstrates that the asymmetry persists even after structural confounds are eliminated.

Finally, the synthetic setting is intentionally simple. It strips away all semantic structure and therefore cannot speak to reasoning, generalization, or world knowledge. Its value lies in its controlled nature: it provides a clean testbed where future work can probe how different architectures, objectives, or optimization regimes handle directionality under tunable entropy.

\subsection{Relation to Prior Work}

The question of whether sequence models exhibit directional preferences has been explored in several natural-language settings. ~\citep{berglund2024reversalcursellmstrained} and subsequent analyses (like ~\citep{lin2024delvingreversalcursefar}) introduced and elaborated on the Reversal Curse, reporting that models trained on facts of the form ``$A$ is the parent of $B$'' often fail to answer the inverse query ``$B$'s parent is what?''. These findings initiated broader examination of whether forward--reverse disparities reflect properties of language itself or inductive biases of autoregressive Transformers.

~\citep{papadopoulos2024arrowstimelargelanguage} directly formulate the problem of forward–reverse asymmetry in sequence modeling. They show that even when a dataset is time-reversed and the same model is trained in both directions, the forward model consistently achieves lower perplexity across languages, architectures, and context lengths. Their analysis frames this discrepancy through the lens of computational irreversibility: many information-preserving operations are easy to compute in the forward direction but significantly harder to invert. In their synthetic constructions, forward prediction resembles a deterministic compression of state (e.g., multiplication), whereas backward prediction requires recovering multiple plausible inputs, akin to factoring. This introduces an inherent complexity asymmetry independent of linguistic priors.

Our benchmark operationalizes this idea in an even more controlled setting. Instead of constructing synthetic languages or relying on natural corpora, we use random string mappings with a tunable branching factor $K$ that explicitly determines the size of the inverse support. Forward mappings ($A\to B$) have zero conditional entropy, while inverse mappings ($B\to A$) have entropy $\log K$. This provides a mathematically precise analogue of the complexity gap described by \citep{papadopoulos2024arrowstimelargelanguage}, but stripped of semantic structure: any residual directional difficulty must arise from architecture, optimization, or initialization rather than statistics of language.

Our findings intersect with research on representation rigidity and pre-training interference. \citep{doi:10.1073/pnas.1611835114} document that reduced plasticity in large, pre-trained models when adapting to distributionally unrelated tasks. Our observation that pre-trained GPT-2 sometimes underperforms random initialization on deterministic synthetic mappings is consistent with this line of work, though our focus is specifically on directionality rather than catastrophic forgetting.

Finally, the capacity limitations observed in our LoRA experiments relate to current discussion around parameter-efficient fine-tuning methods \citep{hu2021loralowrankadaptationlarge, dettmers2023qloraefficientfinetuningquantized}. Prior analyses note that LoRA excels at smoothly modifying pre-trained representations but can struggle with tasks requiring high-entropy memorization or extensive support recovery. Our results may be viewed as an extreme instance of this behavior, in a regime where the mapping is deliberately devoid of semantic structure.

Overall, this benchmark does not challenge or contradict natural-language studies; rather, it provides a controlled environment that captures a minimal core of the asymmetry they report. By quantifying the information-theoretic difficulty and enforcing symmetry in the data support, we offer a framework for attributing directional effects to architecture, optimization, or initialization---independently of linguistic priors.

\section{Limitations}

This benchmark intentionally trades realism for control, introducing several constraints:

\begin{itemize}[nosep]
\item \textbf{Synthetic Scope.}  
The data consists of uniformly random strings with no semantic, grammatical, or hierarchical structure. The results therefore characterize optimization dynamics of the architecture rather than linguistic competence or semantic reasoning. Extrapolating findings to natural language tasks is inappropriate.

\item \textbf{Model Scale.}  
Experiments use GPT-2 Small ($\approx 124$M parameters) as a representative causal Transformer. While the directional gaps are robust across seeds and $K$ values at this scale, larger models may exhibit different degrees of reversibility or optimization efficiency. We do not claim scale laws for directional asymmetry.

\item \textbf{Tokenizer Mismatch.}  
A BPE tokenizer trained on English text is applied to random character sequences. This produces fragmented subword segmentations that slightly inflate the effective input length. Because both directions use identical tokenization, the effect is symmetric, but it does reduce comparability to natural-language workloads.

\item \textbf{Training Configuration.}  
We evaluate under a fixed, standard training regime (AdamW, batch size, learning rate). High-entropy inverse tasks are sensitive to optimization hyperparameters; alternative schedules might shift absolute excess-loss values but are unlikely to reverse the observed qualitative hierarchy between architectures.

\end{itemize}

Overall, the benchmark is best interpreted as a diagnostic tool for architectural inductive biases under controlled entropy, not as a predictor of performance on semantic, reasoning-heavy, or real-world language tasks.

\section{Conclusion}

We introduced a fully synthetic, entropy-controlled benchmark for isolating directional optimization behavior in sequence models. By pairing random string mappings with analytically determined loss floors, the benchmark exposes inefficiencies that cannot be attributed to semantics, token statistics, or corpus-level asymmetries.

Across this controlled setting, three findings consistently emerge. First, scratch-trained Transformers exhibit a pronounced and reproducible directional gap: inverse mappings with nonzero conditional entropy incur substantially higher excess loss than their deterministic forward counterparts, even though both directions are trained under symmetric prompts and identical data support. Second, this asymmetry is not an inherent property of the inverse task itself; a non-causal MLP trained on the same mappings shows only a minor gap, indicating that the effect is tied to the causal Transformer architecture or its optimization dynamics. Third, pre-trained initializations and low-rank adaptation modify optimization efficiency in predictable ways—pre-training introduces rigidity on deterministic mappings, and LoRA encounters a sharp capacity bottleneck on high-entropy inverse tasks—but neither eliminates the core directional effect observed in scratch Transformers.

Because the benchmark eliminates all linguistic priors, these results reveal a minimal, semantics-free signature of directional friction intrinsic to causal Transformer training. The persistence of this effect, even in a uniform, structureless environment, suggests that directional inefficiency is not solely a byproduct of natural-language corpora but reflects deeper architectural or optimization-level biases.

This benchmark is intentionally simple: it does not capture reasoning, generalization, or semantic structure. Its value lies in providing a clean, quantifiable substrate on which future work can dissect the mechanisms underlying reversibility, gradient flow, support recovery, and entropy-sensitive optimization in modern sequence models.

\section{Impact Statement}
This work presents a controlled synthetic benchmark for studying directional training asymmetries in sequence models, with a focus on isolating architectural and optimization effects from data confounds. While the benchmark advances our understanding of inductive biases in Transformers and could inform future model designs to mitigate such asymmetries, it is important to acknowledge its limitations in broader societal contexts.
As a deliberately narrow, toy-scale study using random strings devoid of real-world semantics, the results should not be overgeneralized to natural language applications or used to ``debunk'' phenomena like the Reversal Curse in deployed LLMs. Misuse could arise if the findings are extrapolated without caveats, potentially misleading practitioners or policymakers about the persistence of directional biases in semantically rich, large-scale settings. For instance, claiming that architectural reversal invariance alone resolves real-world failures ignores the interplay with linguistic priors, corpus statistics, and scale---factors explicitly excluded here.
Ethically, this research poses minimal risks: it involves no human subjects, sensitive data, or direct applications to high-stakes domains like healthcare or finance. However, we encourage responsible citation and interpretation, emphasizing the benchmark's role as a diagnostic tool rather than a comprehensive explanation. Future extensions could explore impacts on fairness or robustness in more realistic tasks, but no immediate negative societal consequences are anticipated from this foundational analysis.

\section*{Acknowledgments}
We thank colleagues and reviewers for their constructive feedback and discussions that helped clarify the scope of this work. Any errors are our own.

\bibliographystyle{plainnat}

\bibliography{final}

\onecolumn

\appendix
\section{Implementation Details}
\label{app:implementation}

This appendix summarizes the concrete implementation of the experimental harness used for all reported Transformer results.

\subsection{Data Generation and Topology}

All experiments operate on a synthetic alphabet
\[
\Sigma = \{\texttt{a},\dots,\texttt{z},\texttt{0},\dots,\texttt{9}\}, \quad |\Sigma| = 36,
\]
and fixed-length strings of length $L=8$. For each branching factor $K \in \{1,5,8\}$ and total pair count $n_{\text{pairs}}$, we generate a list of pairs $(A,B)$ using a single deterministic procedure per $K$:
\begin{itemize}[nosep]
    \item For $K = 1$ (bijective regime), we sample $n_{\text{pairs}}$ unique inputs $A$ and $n_{\text{pairs}}$ unique targets $B$ uniformly from $\Sigma^L$, then apply a random permutation to construct a one-to-one mapping.
    \item For $K > 1$ (many-to-one regime), we enforce $K \mid n_{\text{pairs}}$ and set $n_{\text{targets}} = n_{\text{pairs}} / K$. We then sample $n_{\text{targets}}$ unique targets $B_1,\dots,B_{n_{\text{targets}}}$ and, for each $B_i$, sample $K$ distinct inputs $A_{i,1},\dots,A_{i,K}$, yielding $K$ pairs that share the same $B_i$. The list of pairs is finally shuffled.
\end{itemize}

The script logs, for each $K$, the number of unique inputs and targets and verifies that for $K>1$ the number of distinct $B$ matches the expected $n_{\text{pairs}} / K$. No train/validation/test split is used in this version; all pairs are used as training data to isolate optimization dynamics rather than generalization behavior.

\subsection{Prompt Formatting and Label Masking}

Models are trained in a standard causal language modeling setup using HuggingFace \texttt{AutoModelForCausalLM} and \texttt{AutoTokenizer}. For each pair $(A,B)$, we construct a textual prompt and target sequence as
\begin{align*}
    \text{Forward } (A \to B): \quad & \texttt{"x: A y:"} \;\;\to\;\; \texttt{"x: A y: B"} \\
    \text{Backward } (B \to A): \quad & \texttt{"x: B y:"} \;\;\to\;\; \texttt{"x: B y: A"}.
\end{align*}
The tokenizer is loaded from the specified \texttt{model\_id} (GPT-2 in our experiments); if no pad token is defined, the EOS token is reused as the pad token. Inputs are tokenized with truncation and padding to a fixed maximum length (default $\text{max\_len}=64$).

For each example, we construct \texttt{labels} by copying the tokenized sequence and then:
\begin{itemize}[nosep]
    \item setting the label positions corresponding to the prompt prefix (\texttt{"x: \{src\} y:"}) to \texttt{-100}, and
    \item setting labels at padding positions to \texttt{-100}.
\end{itemize}
This yields a symmetric supervision scheme in which the loss is computed only over the target span (either $B$ or $A$), while the prompt tokens act purely as conditioning context.

\subsection{Training Regimes and Hyperparameters}

The script supports four training regimes, selected via \texttt{--modes}:
\begin{enumerate}[nosep]
    \item \textbf{Scratch}: GPT-2 Small is instantiated from configuration (\texttt{from\_config}) with random initialization. This regime isolates the inductive bias of the architecture without any pre-training.
    \item \textbf{Finetune}: GPT-2 Small is loaded from pretrained weights (\texttt{from\_pretrained}) and all parameters are updated. This tests how fully pre-trained representations adapt to the synthetic mapping.
    \item \textbf{Finetune\_Reg}: As in \texttt{Finetune}, but with elevated dropout rates injected into the configuration (\texttt{attn\_pdrop}, \texttt{resid\_pdrop}, \texttt{embd\_pdrop}) and larger weight decay (default \texttt{reg\_weight\_decay = 0.1}). This provides a regularized finetuning baseline.
    \item \textbf{LoRA}: GPT-2 Small is loaded from pretrained weights and wrapped with a low-rank adaptation using \texttt{peft.LoraConfig}, applied to attention projection modules (\texttt{c\_attn}, \texttt{c\_proj}). Only the LoRA parameters are trainable.
\end{enumerate}

For each regime, all runs for a given $K$ share the exact same synthetic pair list, ensuring that architectural comparisons are made under identical data manifolds. Training uses:
\begin{itemize}[nosep]
    \item AdamW optimizer with weight decay $=0.01$ (except \texttt{Finetune\_Reg}, which uses \texttt{reg\_weight\_decay}),
    \item base learning rate specified by \texttt{--lr} (default $10^{-4}$) and fixed across non-LoRA regimes,
    \item gradient clipping with norm $1.0$,
    \item a linear learning-rate schedule with $10\%$ warmup steps, computed as a fraction of the total training iterations.
\end{itemize}

For LoRA, the script sweeps over a list of ranks (\texttt{--lora\_ranks}, default $\{8,64,256\}$). The effective learning rate for each rank is scaled as
\[
\eta_r = \min\left(\eta_{\text{base}} \cdot \frac{r}{8}, 10^{-3}\right),
\]
with fixed weight decay $=0.01$ and LoRA dropout $=0.05$.

\subsection{Excess Loss and Logging}

For each run, we compute the theoretical cross-entropy floor:
\[
\mathcal{L}_{\min} =
\begin{cases}
0, & \text{for } A \to B \\
\log K, & \text{for } B \to A \text{ with } K>1,
\end{cases}
\]
and record the \emph{Excess Loss} as $\mathcal{L}_{\text{excess}} = \mathcal{L}_{\text{train}} - \mathcal{L}_{\min}$ using natural logarithms. Training is performed for a fixed number of epochs (default \texttt{--epochs = 20}); at each epoch the script logs the average training loss, and every few epochs it reports both the raw loss and the excess loss relative to the floor.

At the end of each run, the script logs:
\begin{itemize}[nosep]
    \item the final training loss and excess loss,
    \item runtime (wall-clock seconds),
    \item total and trainable parameter counts,
    \item LoRA-specific trainable parameter counts (where applicable),
    \item the full training loss curve across epochs.
\end{itemize}
All metadata and metrics are serialized as JSON lines into the specified output file, enabling post-hoc aggregation into the summary tables reported in the main text.

\subsection{Reproducibility}

Determinism is enforced as follows:
\begin{itemize}[nosep]
    \item a fixed seed is used per $(K,\text{mode})$ configuration for Python's \texttt{random}, NumPy, and PyTorch RNGs;
    \item \texttt{torch.backends.cudnn.deterministic} is set to \texttt{True} and \texttt{benchmark} to \texttt{False};
    \item the synthetic pair list for a given $K$ is generated exactly once and reused across all regimes and directions.
\end{itemize}
Together, these choices ensure that re-running the script with the same arguments reproduces the same data, loss curves, and excess-loss statistics up to floating-point and hardware variation.

\section{Additional Experimental Figures}

This appendix presents supplementary visualizations corresponding to the results
reported in Sections~3 and~4. All figures are derived directly from the JSONL
logs generated by the public code release.

\subsection{Optimization Friction and Directional Gaps}

Figure~\ref{fig:directional_friction} reports the Excess Loss for forward and inverse
tasks for both the MLP and GPT-2 (scratch) models at $K \in \{5, 8\}$. These
plots illustrate the core phenomenon measured by the benchmark: even under fully
synthetic, entropy-controlled conditions, the causal Transformer exhibits a larger
directional efficiency gap than the non-causal MLP baseline.

\subsection{Pre-training Effects: The Plasticity Tax}

\begin{figure}[h]
    \centering
    \includegraphics[width=0.85\linewidth]{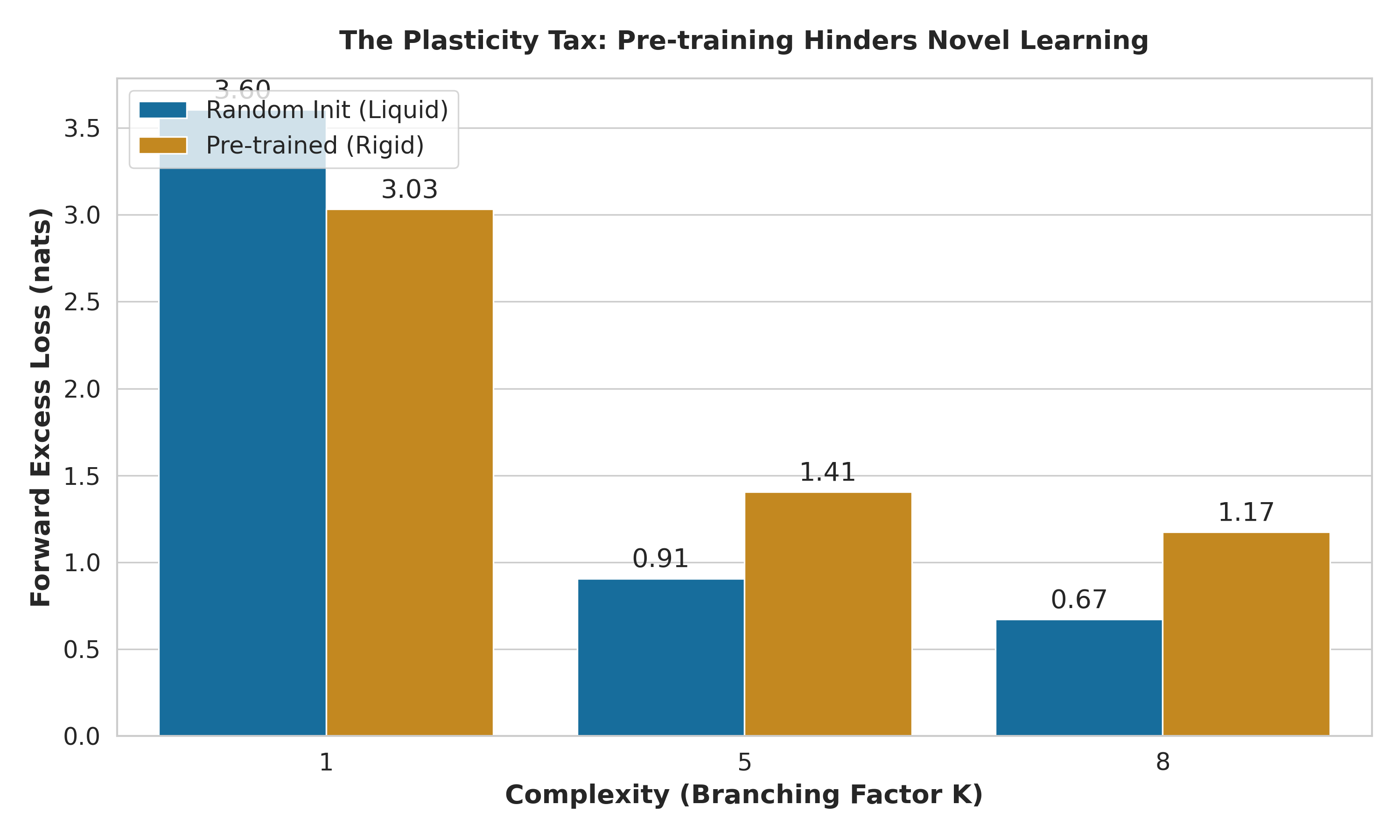}
    \caption{\textbf{The Plasticity Tax.} Pre-trained weights reduce efficiency
    on synthetic deterministic mappings, yielding higher forward excess loss
    relative to random initialization.}
    \label{fig:plasticity_tax}
\end{figure}

Figure~\ref{fig:plasticity_tax} compares Excess Loss for scratch vs.\ pre-trained
initializations across branching factors. While pre-training typically improves
downstream language tasks, here it introduces additional resistance when learning
arbitrary deterministic mappings, yielding higher forward excess loss.

\subsection{Low-Rank Adaptation: Capacity Limits}

\begin{figure}[h]
    \centering
    \includegraphics[width=0.85\linewidth]{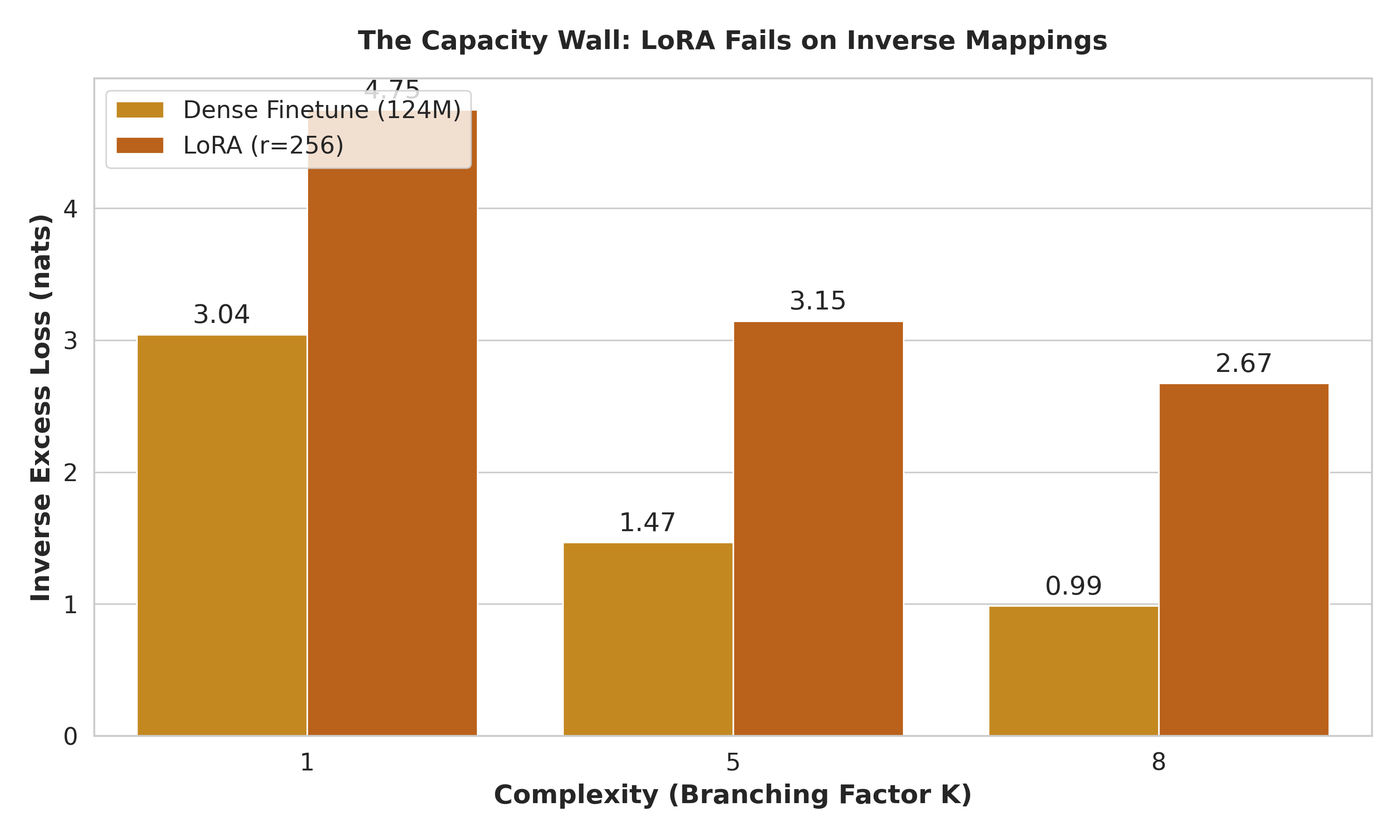}
    \caption{\textbf{LoRA capacity wall.} Even high-rank LoRA struggles to match
    dense fine-tuning on inverse tasks, indicating limited expressivity for
    arbitrary high-entropy mappings.}
    \label{fig:lora_capacity}
\end{figure}

Figure~\ref{fig:lora_capacity} examines the behavior of LoRA (rank $r=256$)
relative to full dense fine-tuning. The inverse ($B\!\to\!A$) task exhibits a
sharp degradation in convergence under low-rank adaptation, consistent with
capacity bottlenecks when fitting high-entropy inverse mappings.

\subsection{Training Dynamics for Inverse Tasks}

\begin{figure}[h]
    \centering
    \includegraphics[width=0.85\linewidth]{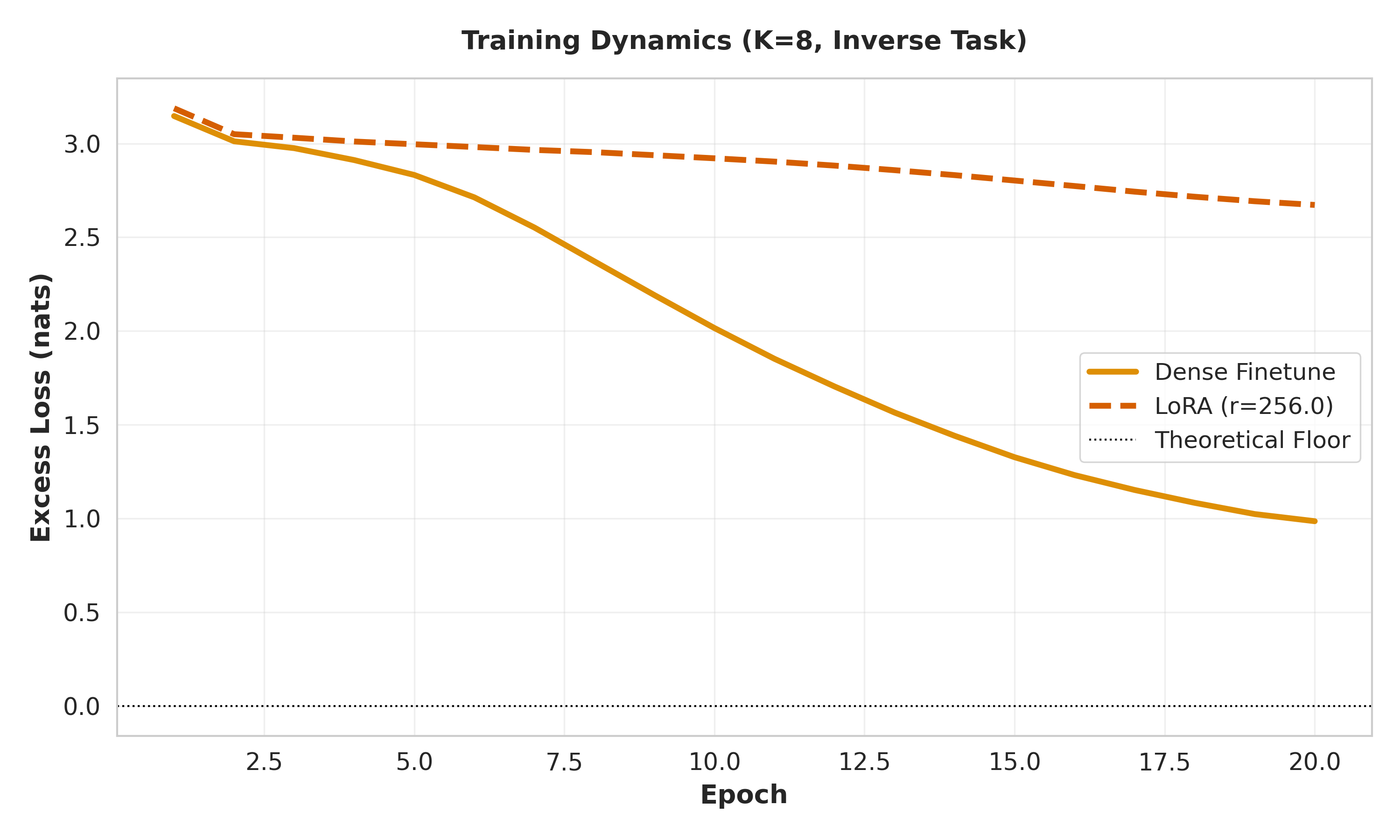}
    \caption{\textbf{Training dynamics on $B\!\to\!A$ (K=8).} Dense fine-tuning
    approaches the theoretical floor; LoRA plateaus early, consistent with a
    rank-limited bottleneck.}
    \label{fig:inverse_dynamics}
\end{figure}

Figure~\ref{fig:inverse_dynamics} presents epoch-wise training curves for LoRA
and dense fine-tuning at $K=8$ on the inverse task. Dense training steadily
approaches the entropy floor, while LoRA plateaus early, showing minimal progress.

\clearpage

\end{document}